\documentclass{article}

\usepackage{arxiv}

\usepackage[utf8]{inputenc} 
\usepackage[T1]{fontenc}    
\usepackage{hyperref}       
\usepackage{url}            
\usepackage{booktabs}       
\usepackage{amsfonts}       
\usepackage{nicefrac}       
\usepackage{microtype}      
\usepackage{lipsum}		
\usepackage{graphicx}
\usepackage{natbib}
\usepackage{doi}

\usepackage{multirow}
\usepackage{amsmath}

\title{Affine transformation estimation improves visual self-supervised learning}

\author{ \href{https://orcid.org/0000-0003-2822-7146}{\includegraphics[scale=0.06]{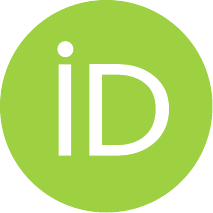}\hspace{1mm}David ~Torpey}\thanks{Alternative email address: 674425@students.wits.ac.za.} \\
	School of Computer Science and Applied Mathematics\\
	University of the Witwatersrand, Johannesburg\\
	South Africa \\
	\texttt{torpey.david93@gmail.com} \\
	\And
	\href{https://orcid.org/0000-0003-0783-2072}{\includegraphics[scale=0.06]{orcid.pdf}\hspace{1mm}Richard ~Klein} \\
	School of Computer Science and Applied Mathematics\\
	University of the Witwatersrand, Johannesburg\\
	South Africa \\
	\texttt{kleinric@gmail.com} \\
}

\renewcommand{\shorttitle}{Affine transformation estimation improves visual SSL}

\hypersetup{
pdftitle={Affine transformation estimation improves visual self-supervised learning},
pdfsubject={cs.CV},
pdfauthor={David ~Torpey, Richard ~Klein},
pdfkeywords={Deep learning, Self-supervised learning, Representation learning, Affine transformation},
}

\begin{document}
\maketitle

\begin{abstract}
The standard approach to modern self-supervised learning is to generate random views through data augmentations and minimise a loss computed from the representations of these views. This inherently encourages invariance to the transformations that comprise the data augmentation function. In this work, we show that adding a module to constrain the representations to be predictive of an affine transformation improves the performance and efficiency of the learning process. The module is agnostic to the base self-supervised model and manifests in the form of an additional loss term that encourages an aggregation of the encoder representations to be predictive of an affine transformation applied to the input images. We perform experiments in various modern self-supervised models and see a performance improvement in all cases. Further, we perform an ablation study on the components of the affine transformation to understand which of them is affecting performance the most, as well as on key architectural design decisions.
\end{abstract}


\section{Introduction}
The amount of data, of all modalities, is increasing at an exponential rate. The vast majority of this data is unlabelled. The process of labelling is a bottleneck because it is both time-consuming and expensive. Such an environment makes algorithms that only require unlabelled data particularly useful and important. This is partially the reason that so-called self-supervised learning (SSL) techniques are such an active area of research in computer vision \citep{simclrv2,barlow_twins,mocov2,dino}. These are a class of unsupervised algorithms that derive a supervision signal from the data itself.

Many types of self-supervised algorithms exist, however, the majority of the popular and successful ones adopt a multi-view perspective \citep{multiviewssl}. This is where two `views` are created using random data augmentation, and a contrastive or predictive learning objective is optimised. Due to the pervasiveness and success of this multi-view approach, we propose a module that naturally fits into this paradigm. Even with their success, SSL techniques are still very computationally expensive to train \citep{simclrv2,byol}, and thus we aim to provide a general technique to improve both the downstream performance and the convergence rate of these models.

The intuition behind our proposed affine module is that any system tasked with understanding images can benefit from understanding the geometry of the image and the objects within it. An affine transformation is a geometric transformation that preserves parallelism of lines. It can be composed of any sequence of rotation, translation, shearing, and scaling. Mathematically, an affine transformation is shown in Equation \ref{eq:affine_transformation}. It has 6 degrees of freedom and is applied to a vector in homogenous coordinates.

\begin{equation}
\label{eq:affine_transformation}
H_\phi = \left[\begin{array}{ccc}
            \phi_{1,1} & \phi_{1,2} & \phi_{1,3}\\
            \phi_{2,1} & \phi_{2,2} & \phi_{2,3}\\
            0 & 0 & 1\\
\end{array}\right]
\end{equation}

The ability to know how a source image was transformed to get to a target image implicitly means that you have learned something about the geometry of that image. An affine transformation is a natural way to encode this idea. Forcing the network to estimate the parameters of a random affine transformation applied to the source images thereby forces it to learn semantics about the geometry. This geometric information can supplement the signal provided by the loss of the underlying SSL technique into which our module is embedded. In this way, we enable the representations to remain invariant to the transformation from the SSL-based loss, while encouraging the representations to be equivariant to (i.e. predictive of) an arbitrary affine transformation from our affine module's loss.

In this paper, we introduce a model-agnostic module that can be plugged into any multi-view-based self-supervised algorithm to improve performance by augmenting the SSL supervision signal with the implicit geometric information encoded in our affine-based loss. The architecture of our proposed approach can be seen in Figure \ref{fig:arch_diagram}. Our affine module essentially manifests as additional loss terms that regress the parameters of an affine transformation applied to the source image views from a given latent \emph{transition vector}. We conduct various analyses and ablations of the important components of our pipeline. We analyse the representation aggregation strategy used to compute the transition vector. We ablate the four constituent transformations of the affine transformation to gauge their relative performance contributions. We analyse whether it is beneficial to perform the affine estimation for both random views. Finally, we analyse the potential source of bias introduced by the background induced after performing an arbitrary affine transformation on an image. Through extensive experiments, we show that the inclusion of our models improves the downstream performance and the convergence time in all studied SSL techniques and downstream datasets, and therefore serves as a helpful additional supervision signal for multi-view SSL. Lastly, we recommend settings for the various important components and hyperparameters of our pipeline that work well in general, while balancing a performance-compute tradeoff.

The remainder of the paper is structured as follows. In Section \ref{sec:rw}, we cover the related work in the area of self-supervised learning, going into detail where necessary. In Section \ref{sec:method} we detail our proposed method. We then delve into the details behind the architecture and choices for the various parts of the system. This is followed by a comprehensive set of experiments in Section \ref{sec:exp}, including results of various datasets, as well as an ablative study. Finally, the paper is concluded with some closing remarks in Section \ref{sec:conclusion}.

\section{Related Work}
\label{sec:rw}
SSL can largely be separated into two main paradigms: instance discrimination and pretext tasks. Modern SSL has, to a large extent, focused on instance discrimination methods (the most popular class of methods known as contrastive learning) since they tend to outperform pretext task methods significantly. A pretext task method consists of a manually defined surrogate problem that the network is tasked with solving. The assumption is that if the network can solve this task adequately, then it has learned something useful about the data distribution, and thus the representations it produces are of good quality. Pretext task methods include image jigsaw puzzles \citep{jigsaw}, relative patch prediction \citep{contextpred}, colourisation \citep{colorful_image_colorization}, inpainting \citep{inpainting}, and rotation prediction \citep{rotation_feature_decoupling,predicting_image_rotations}. Pretext task methods typically underperform in comparison to other SSL methods because they are inherently limited in scope due to the requirement of manually defining the task.

Contrastive learning is the most popular class of techniques within instance discrimination and includes models such as SimCLR \citep{simclr,simclrv2}, CPC \citep{cpc}, NPID \citep{npid}, and MoCo \citep{moco,mocov2}. Contrastive techniques rely on some form of contrastive loss, usually the InfoNCE loss, based on NCE \citep{nce}. The high-level goal is to pull similar images together and push dissimilar images apart in latent space. Other techniques that derive a supervision signal from a loss based on the representations are various non-contrastive methods such as BYOL \citep{byol}, SwAV \citep{swav}, SimSiam \citep{simsiam}, DINO \citep{dino}, and Barlow Twins \citep{barlow_twins}. These methods consist of a wide variety of approaches, including clustering \citep{swav}, redundancy reduction \citep{barlow_twins}, and teacher-student self-distillation \citep{byol}.

\begin{figure}[!h]
    \centering
    \includegraphics[scale=0.2]{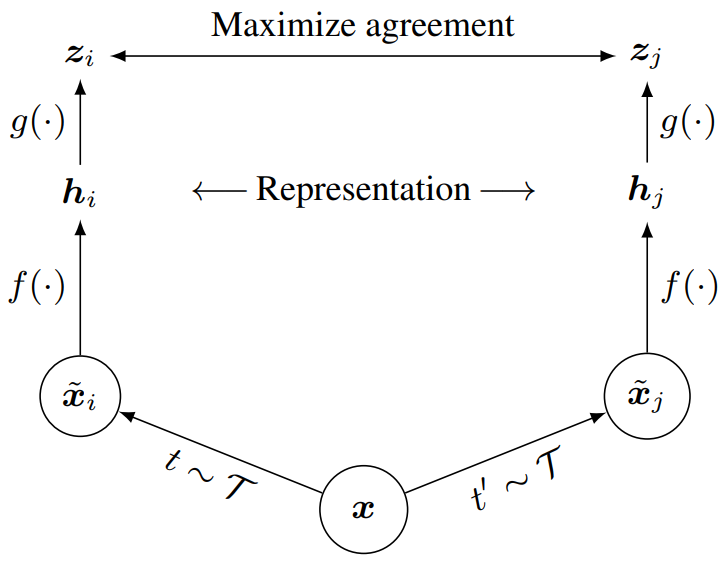}
    \caption{SimCLR \citep{simclr} architecture.}
    \label{fig:simclr_arch}
\end{figure}

Since we choose SimCLR, BYOL, and Barlow Twins as the SSL methods to study in this work, we describe them in a little more detail next. The generation of random views is a fundamentally important process in modern SSL. To generate random views, SimCLR uses a random data augmentation pipeline consisting of various transformations such as random cropping, colour jittering, colour dropping, and Gaussian blurring. An image is sampled from the data, and two random views are generated. These views are sent through an encoder network (usually a ResNet), followed by a projection head to produce latent representations of the views (see Figure \ref{fig:simclr_arch} for SimCLR's architecture). The NT-Xent \citep{simclr} contrastive loss is then computed using a batch of each of these representations and minimised. Once trained, the encoder network can then be used in downstream applications, such as image classification, object detection, and semantic segmentation. Importantly, only the encoder representations are used downstream, and not the projection head representations.

BYOL utilises the same process as SimCLR to generate views, however, the architecture and learning process are different. Firstly, there are two separate networks instead of one: the online and target networks. Similarly, the views are sent through an encoder and projection head. However, the online network contains an additional MLP known as the predictor network (which helps prevent collapsing solutions). The online network is then tasked with predicting the target network's latent representation. The loss is a mean-squared error between the normalised representations from the two networks. Importantly, the online network is updated using backpropagation. The target network is instead updated using an exponential moving average (EMA) of the online network's weights. This EMA procedure ensures consistency of the produced representations during training and helps to avoid collapse.

Barlow Twins is a model inspired by the redundancy-reduction principle. Much of the training procedure is the same as SimCLR: views are generated in the same way (possibly with slightly different base transformations) and a single network is used to map the views to latent representations using an encoder network and projection head MLP. The projection head tends to be much larger in Barlow Twins, as it has empirically shown to require a large capacity in this architecture. The most important difference is the loss function. The network is tasked with making the cross-correlation matrix between the latent representations of the two batches of random views to be as close to the identity matrix as possible. This loss naturally avoids collapse, and much like BYOL, enables Barlow Twins to work well with small batch sizes (unlike SimCLR, which naturally requires a large batch size due to the contrastive loss requiring sufficient negative samples).

\section{Methodology}
\label{sec:method}
\begin{figure}[!h]
    \centering
    \includegraphics[scale=0.4]{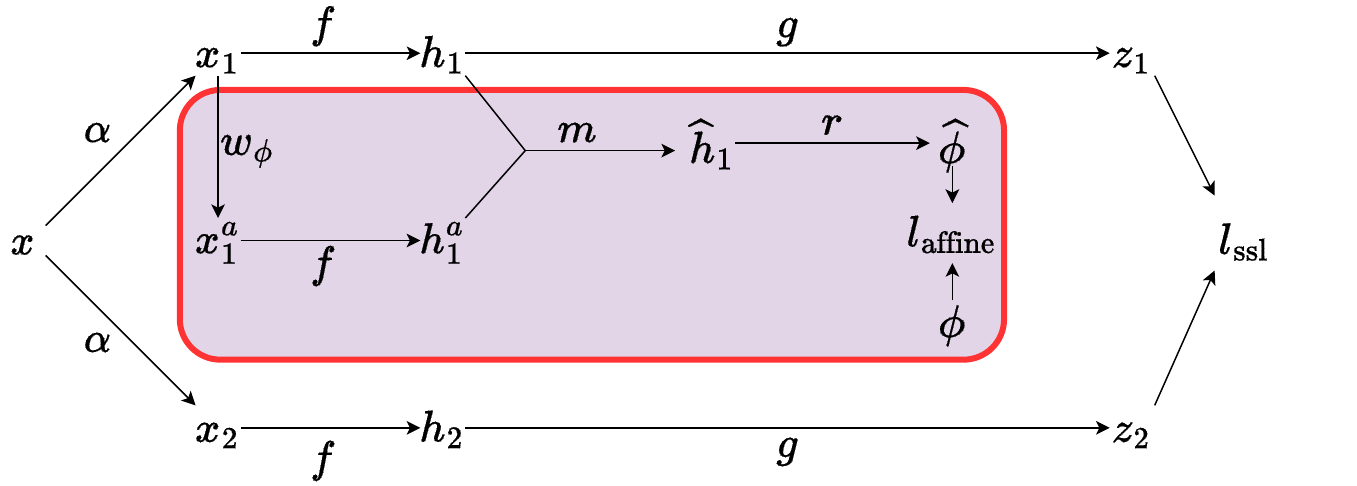}
    \caption{Architecture diagram depicting our method. Our contributions are enclosed in the purple box. The remainder of the diagram illustrates a typical, modern self-supervised method. Note that we only depict the case of one random view being used in our module because the process is exactly the same if the other random view is included.}
    \label{fig:arch_diagram}
\end{figure}

A typical multi-view SSL method will sample an image $x \in X$, where $X$ is the dataset. Two random views will then be created using a stochastic augmentation function $\alpha \in \mathcal{A}$: $x_1, x_2 = \alpha(x)$, for some (possibly infinite) set of augmentations $\mathcal{A}$. These two views are necessarily different due to the stochasticity of $\alpha$, and each element of $\mathcal{A}$ is a function that applies a pipeline of randomly chosen transformations from a set of base transformations such as cropping, blurring, and colour jitter. The base transformations we use are those from the literature \citep{simclr}, but can be any arbitrary set of transformations. Next, $x_1$ and $x_2$ are projected into a latent space using an encoder $f$; $h_1 = f(x_1)$ and $h_2 = f(x_2)$ (note that two different encoders can operate on $x_1$ and $x_2$, but we consider the case where the weights are shared -- without loss of generality). A projection head $g$ -- defined by a multi-layer perceptron (MLP) -- is then used to project these representations into a final latent space, yielding $z_1$ and $z_2$. The network weights are then updated using the gradients from a loss of the form $l_{\texttt{ssl}} = L(z_1, z_2)$. It is clear that such a loss will encourage the network to be invariant to $\alpha$, and thus, the transformation that it is comprised of.

We propose a module that is tasked with estimating an affine transformation of the input random views (depicted in Figure \ref{fig:arch_diagram}). We parameterise this module as an MLP $r$, and compute the affine transformations using a function $w_{\phi}$, where $\phi$ is the parameter vector of the applied affine transformation. The parameter vector is of the following form:
\begin{equation}
    \phi = [\theta, t_x, t_y, \sigma, s_x, s_y]
\end{equation}
where $\theta$ is the rotation angle, $t_x$ and $t_y$ are the horizontal and vertical translation, $\sigma$ is the scale, and $s_x$ and $s_y$ are the horizontal and vertical shear angles.

To estimate the affine transformation, we first compute the affine transformation of input view $x_1$: $x_1^a = w_{\phi_1}(x_1)$. Next, we compute the latent representation in the same way as the random views: $h_1^a = f(x_1^a)$. Note that we only compute the encoder representation for the affine-transformed views and not the projection head representation. In this way, $f$ will learn to produce representations sensitive to affine changes, while $g$ will learn to be invariant to transformations within $\mathcal{A}$.

The vectors $h_1$ and $h_1^a$ are then aggregated using an aggregation function $m$: $\hat{h}_1 = m(h_1, h_1^a)$. We term $\hat{h}_1$ the latent \emph{transition vector}, since it encodes the affine transformation (i.e. how to transition from the original random view to the affine-transformed version). We test various options for $m$ in our experiments, including vector difference and concatenation. Finally, the ground truth affine transformation parameter vector is estimated: $\hat{\phi_1} = r(\hat{h}_1)$. The loss of the module is then defined as the mean squared error between the estimated and ground truth parameter vectors: $l_{\texttt{affine}} = \texttt{MSE}(\phi_1, \hat{\phi_1})$.

The entire loss is then defined as:
\begin{equation}
    \label{eqn:full_loss}
    l = \beta_1 l_{\texttt{ssl}} + \beta_2 l_{\texttt{affine}}
\end{equation}
where $\beta_1$ and $\beta_2$ control the weight given to the SSL and affine loss terms, respectively. It should be noted that we experiment with adding a loss term for $\texttt{MSE}(\phi_2, \hat{\phi_2})$, which is computed by performing the same process for input view $x_2$. By casting the affine objective in this way, instead of encouraging invariance as in the SSL loss $l_{\texttt{ssl}}$, we are encouraging the encoder representation to be predictive of an affine transformation.

Importantly, the additional overhead of our method in terms of the number of parameters that require estimation is simply the number of parameters in $r$. This additional module only accounts for approximately $4\%$ of the total number of parameters in the model. Further, it should be noted where our work differs from the existing work from the literature. In particular, previous work of a similar nature contains methods that solely focus on the prediction of image rotations \citep{rotation_feature_decoupling,predicting_image_rotations}. \cite{predicting_image_rotations} task a network with predicting the rotation angle ($0^{\circ}$, $90^{\circ}$, $180^{\circ}$, or $270^{\circ}$), casting it as a four-way classification problem. \cite{rotation_feature_decoupling} decouple the learning of rotation invariance from the discrimination of individual instances. The rotation angle is predicted in the same way as \cite{predicting_image_rotations} (i.e. as a classification problem), while a contrastive loss is used to discriminate instances in order to decouple the rotation-irrelevant information from the contrastive prediction. However, these methods are only superficially related to our method, as we propose a novel, model-agnostic module that can be used to improve the performance of multi-view SSL algorithms. Further, we focus on a generalised geometric transformation (affine), instead of solely the prediction of four rotation angles.

\subsection{An Invariance Perspective}
We briefly discuss how our method relates to encoding transformation invariance and/or sensitivity into a multi-view SSL architecture. For encoder $f$ to be invariant to a transformation $\alpha \in \mathcal{A}$, we must have that $f(x) = f(\alpha(x)) \texttt{  } \forall x$, where $\mathcal{A}$ is a set of possible transformations. Since $f$ is learned, we can encourage invariance in $f$ by minimising $f(x) - f(\alpha(x))$, which can be expressed as minimising $L(f(x), f(\alpha(x)))$ for some loss function $L$. Further, we can rewrite our loss from Eqn. \ref{eqn:full_loss} as:
\begin{equation}
\label{eqn:expanded_loss}
\begin{split}
    l = \beta_1 \mathbb{E}[L(g(f(\alpha_1(x))), g(f(\alpha_2(x))))] + \\
    \beta_2 \mathbb{E}[\texttt{MSE}(\phi, r(m(f(\alpha_1(x)), f(w_{\phi}(\alpha_1(x)))))]
\end{split}
\end{equation}
where $\alpha_1, \alpha_2 \in \mathcal{A}$ are two random augmentation functions (we use this notation here instead of $\alpha$ for brevity, but they represent the same sample space of random data augmentations), and $L$ is some SSL-based loss function. The first term of loss \ref{eqn:expanded_loss} is of the form discussed above that will encourage invariance to the transformations within $\mathcal{A}$. Conversely, the second term of loss \ref{eqn:expanded_loss} is instead formulated to be \emph{predictive} of the affine transformation $w_{\phi}$. Thus, $m$ will learn to produce representations that are sensitive to an arbitrary affine transformation.

In this way, we balance which transformations we are invariant to ($\mathcal{A}$) and which we are equivariant to ($w_{\phi}$). Our method is the case where $\mathcal{A}$ is the common, effective transformation set used to generate random views from the literature \citep{simclr,byol,barlow_twins} (e.g. random crop, horizontal flipping, colour jitter, colour dropping, and Gaussian blurring) and $w_{\phi}$ is an arbitrary affine transformation. We show how balancing invariance and equivariance for this case can greatly improve downstream performance and convergence rate, and conduct a thorough analysis of the various components of such a pipeline.

\section{Experiments}
\label{sec:exp}
We perform various experiments and analyses to comprehensively evaluate and understand our proposed affine transformation estimation module. We look at overall performance when including the module in the learning process of SimCLR, BYOL, and Barlow Twins. Next, we ablate and analyse two key architectural design choices: 1) the number of views to estimate the affine transformation for, and 2) the aggregation strategy parameterised by $m$. Finally, we analyse the effect of the 4 components of an affine transformation on performance. For these experiments, the affine parameter vector would naturally change dimensionality, depending on which of the 4 components are included in it.

\subsection{Experimental Setup}
In all experiments, we pretrain the SSL models on the Tiny ImageNet \citep{tiny_imagenet} dataset at the default resolution. We use a ResNet50 \citep{resnet} backbone for all experiments. Encoder $f$ is parameterised by a ResNet50 and $g$ as an MLP. Note that in the case of BYOL, an additional predictor head is added on top of the projection head of the \emph{online} network. The hyperparameters used for pretraining are detailed in Table \ref{tbl:hyperparams}. Further, we parameterise $r$ as a two-layer MLP with $6$ output nodes, relating to the dimensionality of the affine parameter vector $\phi$. In all cases, we train with cosine learning rate decay and use the standard random augmentation pipeline to generate views \citep{simclr}, namely RandomResizeCrop, HorizontalFlip, ColorJitter, RandomGrayscale, GaussianBlur.

For the affine transformation $w_{\phi}$, we sample, uniformly, the rotation angle $\theta$ from $[-90, 90]$ degrees, the (relative) translations $t_x, t_y$ from $[0, 0.25]$, the scale $\sigma$ from $[0.7, 1.3]$, and the shear values $s_x, s_y$ from $[-25, 25]$ degrees. Further, for the loss, we set $\beta_1 = \beta_2 = 1$ for SimCLR and BYOL, and $\beta_1 = 1$, $\beta_2 = 10$ for Barlow Twins.

\begin{table}[!h]
\caption{Hyperparameters for the pretraining phase for SimCLR, BYOL, and Barlow Twins. Note that `Predictor Head' only applies to BYOL.}
\label{tbl:hyperparams}
\centering
\begin{tabular}{|l|l|l|}
\hline
\multirow{4}{*}{Optimisation} & Learning Rate & 0.03 \\ \cline{2-3} 
 & Weight Decay & 0.0004 \\ \cline{2-3} 
 & Batch Size & 256 \\ \cline{2-3} 
 & Epochs & 100 \\ \hline
\multirow{2}{*}{Projection Head} & Hidden & 512 \\ \cline{2-3} 
 & Output & 128 \\ \hline
\multirow{2}{*}{Predictor Head} & Hidden & 512 \\ \cline{2-3} 
 & Output & 128 \\ \hline
\multirow{2}{*}{Affine Head} & Hidden & 512 \\ \cline{2-3} 
 & Output & 6 \\ \hline
\end{tabular}
\end{table}

We focus on the downstream task of image classification and evaluate across three different datasets: CIFAR10 \citep{cifar10andcifar100}, CIFAR100 \citep{cifar10andcifar100}, and Caltech101 \citep{caltech101}. This will give a variety in our evaluation, and more comprehensively test our proposed affine module. We average all reported performance metrics across 5 random trials. We use the linear evaluation paradigm, which is the standard approach in the literature \citep{simclr,moco,barlow_twins,dino,byol}. We train a multinomial logistic regression model on the representations produced by the backbone ResNet50 encoder, optimised using the L-BFGS algorithm. We use accuracy as the performance metric and provide 95\% confidence intervals for all results.

\subsection{Module Performance}
\begin{table}[!h]
\caption{\label{tbl:main_results}Results of the baseline models and the variants that include the affine module.}
\centering
\begin{tabular}{ll|ccc}
 &  & \multicolumn{1}{l}{CIFAR10} & \multicolumn{1}{l}{CIFAR100} & \multicolumn{1}{l}{Caltech101} \\ \hline
\multirow{2}{*}{SimCLR} & Standard & $52.88 \pm 0.17$ & $26.76 \pm 0.16$ & $58.38 \pm 0.55$ \\
 & + Affine Module & \pmb{$55.1 \pm 0.2$} & \pmb{$28.6 \pm 0.08$} & \pmb{$62.34 \pm 0.4$} \\ \hline \hline
\multirow{2}{*}{BYOL} & Standard & $46.11 \pm 0.21$ & $21.38 \pm 0.33$ & $51.21 \pm 0.56$ \\
 & + Affine Module & \pmb{$52.29 \pm 0.23$} & \pmb{$25.86 \pm 0.38$} & \pmb{$58.45 \pm 0.89$} \\ \hline \hline
\multirow{2}{*}{Barlow Twins} & Standard & $50.31 \pm 0.13$ & $24.48 \pm 0.11$ & $55.08 \pm 0.74$ \\
 & + Affine Module & \pmb{$52.81 \pm 0.33$} & \pmb{$26.23 \pm 0.12$} & \pmb{$58.68 \pm 1.2$} \\ \hline
\end{tabular}
\end{table}

Table \ref{tbl:main_results} shows the results of including the affine module in the 3 SSL models. The inclusion of the model improves performance in all cases, sometimes by a fairly notable margin (note these results are all statistically significant). For example, the module results in a relative increase in performance, when averaged across the 3 downstream datasets, of over $16\%$ for BYOL and approximately $6\%$ for SimCLR and Barlow Twins. 

Interestingly, the performance improvement is most significant for the CIFAR100 dataset for all 3 SSL models. We posit that this is because enforcing the latent representations to encode an affine transformation has a clearer benefit when there is much variety in the data (in terms of classes/objects) since in such a setting there is more room to discriminate between affine views and estimate the associated transformation parameter vector $\phi$. Fewer classes induce less variety in the possible transformations and what the algorithm can learn by discriminating between the random views.

\begin{figure*}[!t]
\minipage{0.32\textwidth}
  \includegraphics[width=\linewidth]{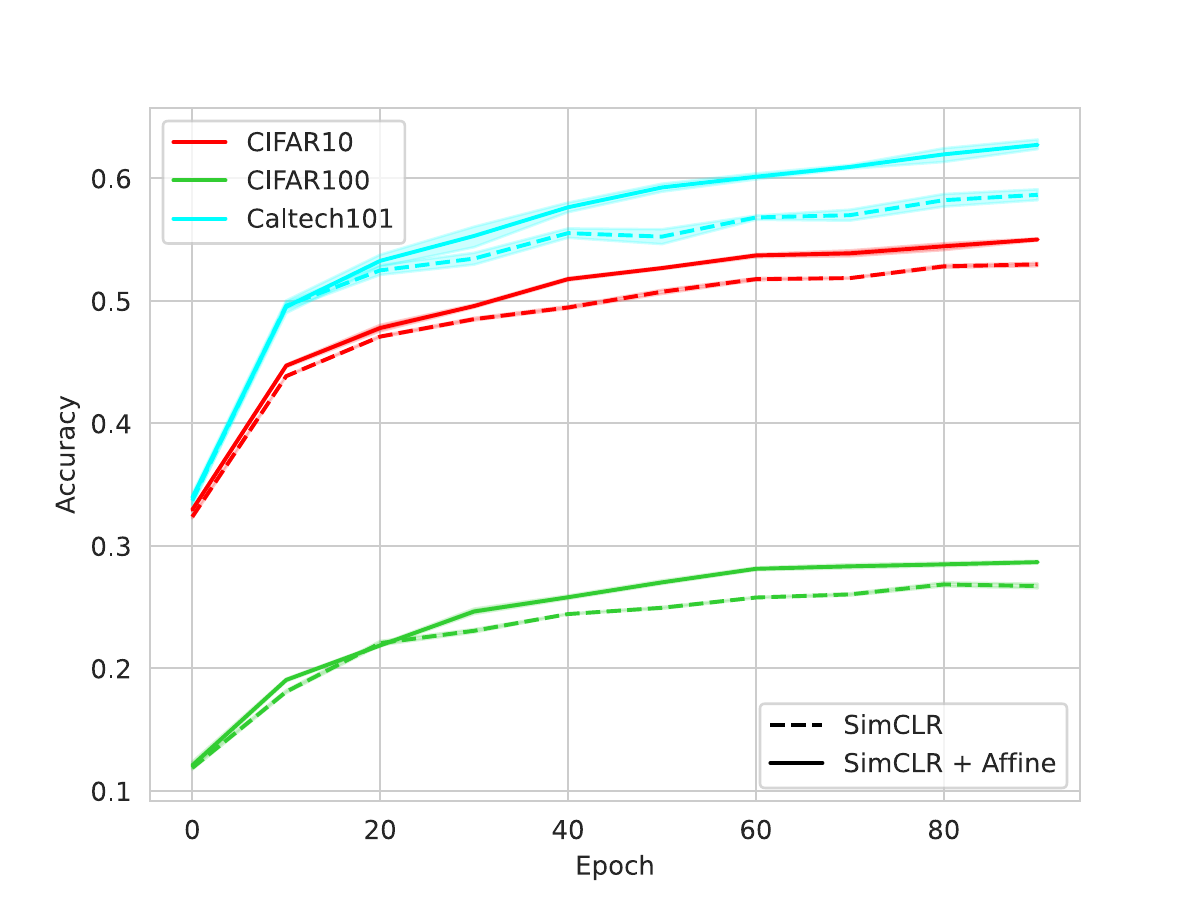}
  \caption{Downstream performance for SimCLR with and without the affine module during training.}\label{fig:simclr_epoch_results}
\endminipage\hfill
\minipage{0.32\textwidth}
  \includegraphics[width=\linewidth]{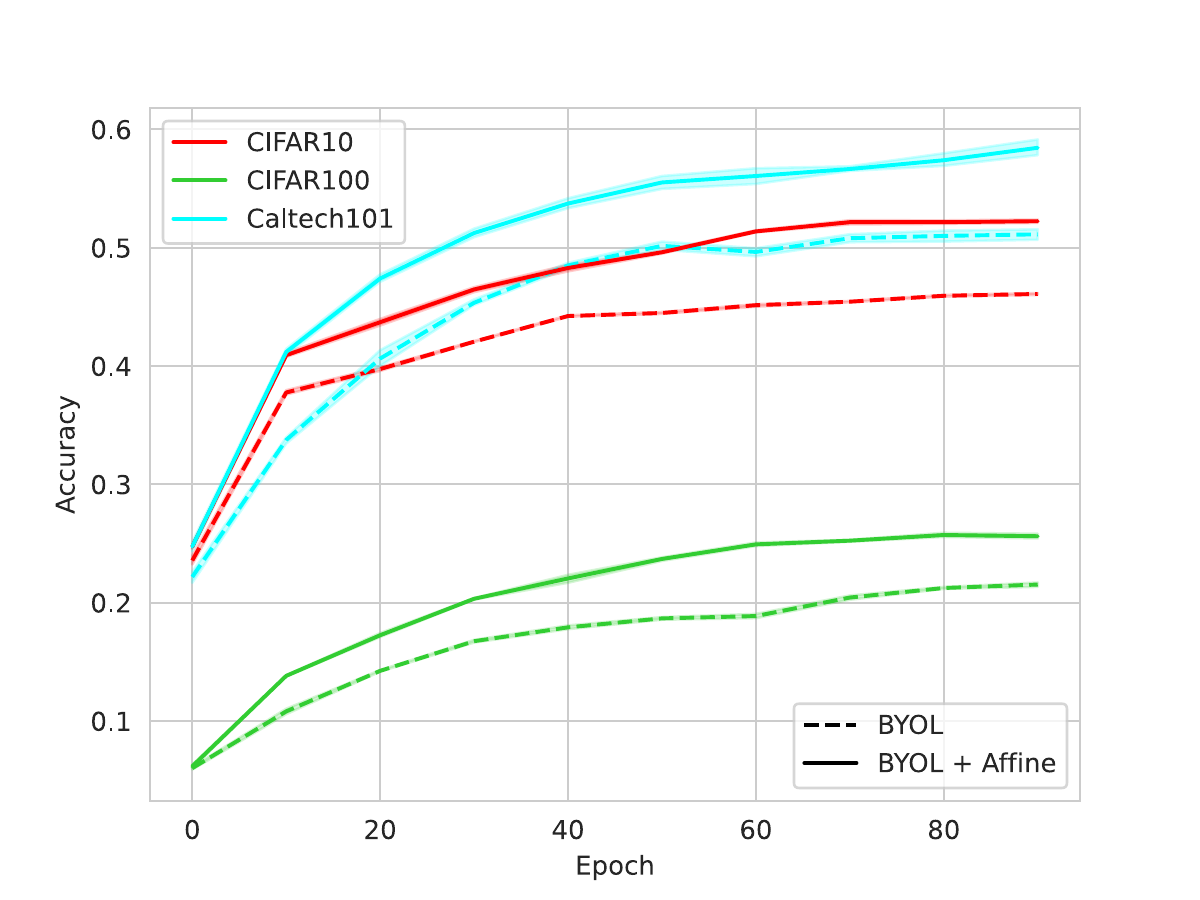}
  \caption{Downstream performance for BYOL with and without the affine module during training.}\label{fig:byol_epoch_results}
\endminipage\hfill
\minipage{0.32\textwidth}%
  \includegraphics[width=\linewidth]{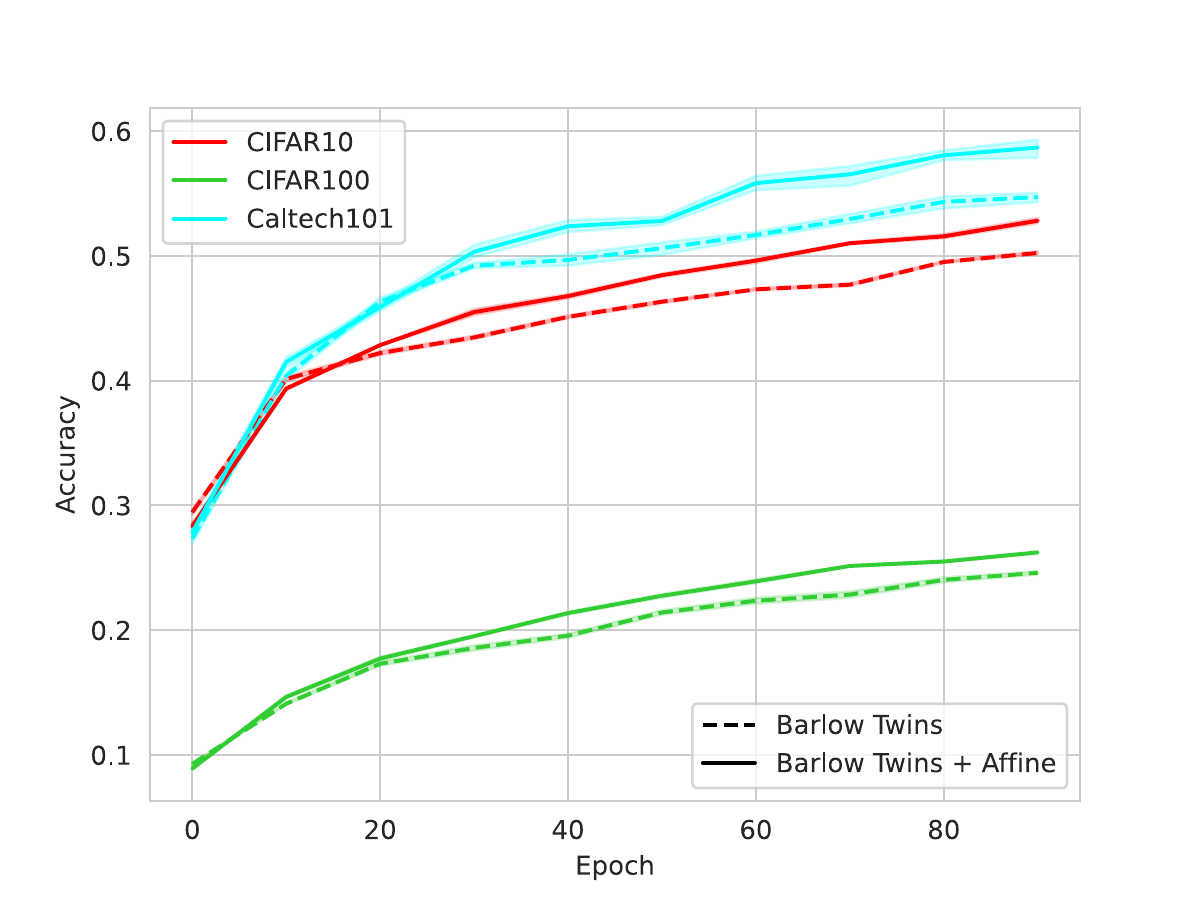}
  \caption{Downstream performance for Barlow Twins with and without the affine module during training.}\label{fig:barlow_twins_epoch_results}
\endminipage
\end{figure*}

Figures \ref{fig:simclr_epoch_results}, \ref{fig:byol_epoch_results}, and \ref{fig:barlow_twins_epoch_results} show the downstream performance of the three SSL methods during training over the pretraining epochs (note that confidence intervals are shown on these three plots, but have low variance). Essentially, we run the evaluation process every 10 training epochs for each model variant and downstream dataset. It is clear from these figures that our affine module improves the performance of all models considerably, particularly as pretraining progresses. We posit that the reason for this performance gap is because $l_{\texttt{affine}}$ is providing complementary information to what $l_{\texttt{ssl}}$ is providing as a supervisory signal to the network. The relative benefit of our module is more evident for BYOL compared to SimCLR and Barlow Twins. This larger benefit for BYOL is likely because the signal provided by our affine module may have less overlap with the information already coming from the BYOL loss (we leave this as an investigation for future work).

\subsection{Architectural Design Choices}
We analyse two important architectural design choices for our model. First, we analyse estimating the affine parameter vector $\phi$ for one random view, or both random views. Next, we analyse different choices of aggregation function $m$: vector difference and concatenation.

\begin{table}[!h]
\caption{\label{tbl:num_views_ablation}Ablation results for performing the affine transformation estimation for one random view ($1 \times$) or both ($2 \times$).}
\centering
\begin{tabular}{ll|rrr}
 & $m$ & \multicolumn{1}{l}{CIFAR10} & \multicolumn{1}{l}{CIFAR100} & \multicolumn{1}{l}{Caltech101} \\ \hline
\multirow{2}{*}{SimCLR + Affine Module} & 1x & \pmb{$55.1 \pm 0.2$} & \pmb{$28.6 \pm 0.08$} & $62.34 \pm 0.4$ \\
 & 2x & $54.33 \pm 0.34$ & $27.7 \pm 0.24$ & \pmb{$62.46 \pm 0.99$} \\ \hline \hline
\multirow{2}{*}{BYOL + Affine Module} & 1x & \pmb{$52.29 \pm 0.23$} & \pmb{$25.86 \pm 0.38$} & \pmb{$58.45 \pm 0.89$} \\
 & 2x & $52.21 \pm 0.27$ & $25.19 \pm 0.28$ & $58.29 \pm 0.92$ \\ \hline \hline
\multirow{2}{*}{Barlow Twins + Affine Module} & 1x & \pmb{$52.81 \pm 0.33$} & \pmb{$26.23 \pm 0.12$} & \pmb{$58.68 \pm 1.2$} \\
 & 2x & $50.86 \pm 0.15$ & $24.36 \pm 0.11$ & $56.99 \pm 0.55$ \\ \hline
\end{tabular}
\end{table}

Table \ref{tbl:num_views_ablation} shows that, in the majority of cases, estimating the affine transformation vector for a single random view performs better than estimating for both random views, sometimes only marginally so. This is not entirely unexpected, since there is a large degree of redundancy in the information provided by the supervision signal derived from the second random view. When the module is applied to both views, the fact that both random views are generated using augmentations from the same sample space $\mathcal{A}$ means that the affine transformation is being applied to roughly the same image distribution, and thus its associated parameters are being estimated for the same distribution. This induces a high degree of redundancy that does not translate to an additional useful supervision signal, and at the cost of additional compute.

\begin{table}[!h]
\caption{\label{tbl:aggregation_strategy_ablation}Ablation results for the aggregation strategy when computing the transition vector during the estimation of the affine transformation (vector difference vs concatenation).}
\centering
\begin{tabular}{ll|ccc}
 &  & \multicolumn{1}{l}{CIFAR10} & \multicolumn{1}{l}{CIFAR100} & \multicolumn{1}{l}{Caltech101} \\ \hline
\multirow{2}{*}{SimCLR + Affine Module} & Diff & \pmb{$55.1 \pm 0.2$} & \pmb{$28.6 \pm 0.08$} & \pmb{$62.34 \pm 0.4$} \\
 & Concat & $53.71 \pm 0.14$ & $27.16 \pm 0.14$ & $62.06 \pm 0.99$ \\ \hline \hline
\multirow{2}{*}{BYOL + Affine Module} & Diff & \pmb{$52.29 \pm 0.23$} & \pmb{$25.86 \pm 0.38$} & \pmb{$58.45 \pm 0.89$} \\
 & Concat & $50.21 \pm 0.21$ & $23.51 \pm 0.25$ & $56.63 \pm 1.19$ \\ \hline \hline
\multirow{2}{*}{Barlow Twins + Affine Module} & Diff & \pmb{$52.81 \pm 0.33$} & \pmb{$26.23 \pm 0.12$} & \pmb{$58.68 \pm 1.2$} \\
 & Concat & $51.74 \pm 0.26$ & $24.94 \pm 0.11$ & $58.37 \pm 0.67$ \\ \hline
\end{tabular}
\end{table}

\begin{figure}[!h]
    \centering
    \includegraphics[scale=0.4]{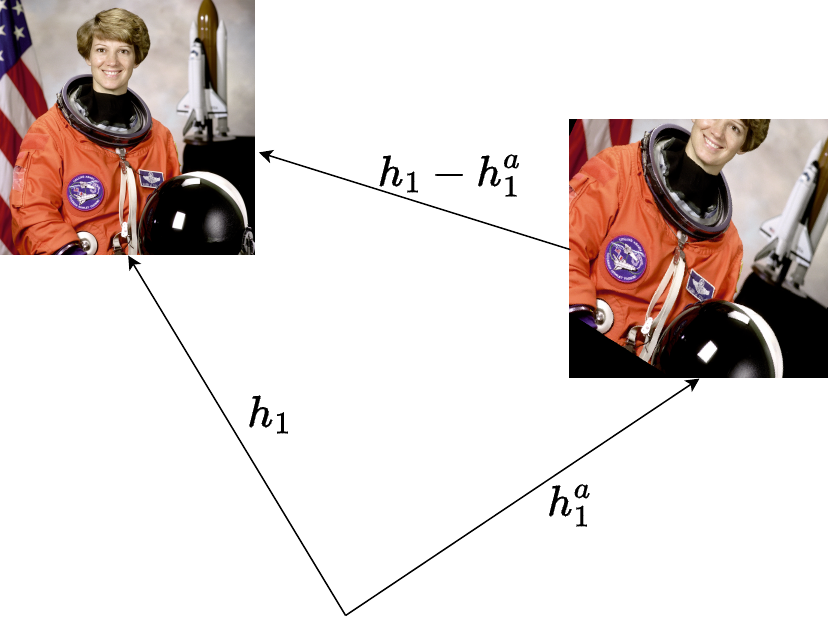}
    \caption{We posit that the reason for $m$ being defined as vector difference resulting in superior performance to concatenation is because the vector difference more naturally encodes the transformation. Concretely, $h_1 - h_1^a$ encodes, in latent space, the vector required to transition from the affine-transformed image back to the original. Conversely, the concatenation of $h_1$ and $h_1^a$ does not encode this transformation in a natural or explicit way. This may in turn make the concatenation harder to estimate the affine parameter vector from.}
    \label{fig:vec_diff_explained}
\end{figure}

Results for different options for aggregation function $m$ are shown in Table \ref{tbl:aggregation_strategy_ablation}. For the majority of models and downstream datasets, vector difference results in superior performance compared to concatenation. This is useful since vector difference reduces the computational complexity of the module by reducing the number of trainable parameters in $r$. We argue that vector difference more naturally encodes the transition vector geometrically than a vector concatenation. We illustrate this further in Figure \ref{fig:vec_diff_explained}. 

\begin{figure}
    \centering
    \includegraphics[scale=0.4]{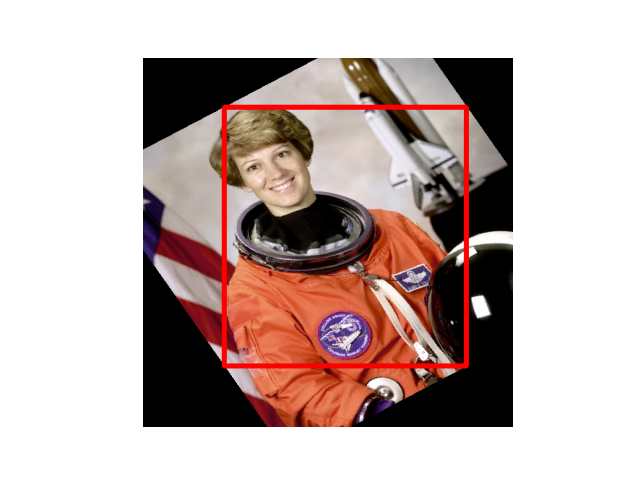}
    \caption{Maximal inscribed axis-aligned rectangle after affine transformation \citep{maxrect}.}
    \label{fig:bounded_process}
\end{figure}

Next, we analyse the effect of the additional background induced during geometric transformations that are not natural parts of the image. These may induce biases during training that potentially affect results. To analyse this, we crop the maximal inscribed axis-aligned rectangle within the region after the affine transformation is performed (see Figure \ref{fig:bounded_process}). This ensures that the induced background is removed while maintaining as much of the transformed image as possible. We leave the rest of the training pipeline the same. Results can be seen in Table \ref{tbl:bounded_results}. Interestingly, in the majority of cases, the default affine transformation performs best, apart from SimCLR in which the bounded transformation results in superior performance for the majority of downstream datasets. It should be noted, however, that the performance difference is negligible in most cases, which suggests that the potential bias from the induced background has little effect on downstream performance. We, therefore, recommend opting for the default affine transformation, since computing the bounded crop introduces a potentially unnecessary computational burden (by solving the optimisation problem for the maximal inscribed rectangle).

\begin{table}[!h]
\caption{\label{tbl:bounded_results}Results for default affine transformation versus the `bounded' version in which the maximal inscribed axis-aligned rectangle is cropped to ensure no induced background is fed into the model during pretraining.}
\centering
\begin{tabular}{ll|rrr}
 &  & \multicolumn{1}{l}{CIFAR10} & \multicolumn{1}{l}{CIFAR100} & \multicolumn{1}{l}{Caltech101} \\ \hline
\multirow{2}{*}{SimCLR} & Default & $55.1 \pm 0.2$ & $28.6 \pm 0.08$ & \pmb{$62.34 \pm 0.4$} \\
 & Bounded & \pmb{$56.02 \pm 0.24$} & \pmb{$29.8 \pm 0.23$} & $61.25 \pm 0.93$ \\ \hline \hline
\multirow{2}{*}{BYOL} & Default & \pmb{$52.29 \pm 0.23$} & \pmb{$25.86 \pm 0.38$} & \pmb{$58.45 \pm 0.89$} \\
 & Bounded & $51.66 \pm 0.23$ & $25.31 \pm 0.2$ & $56.35 \pm 0.83$ \\ \hline \hline
\multirow{2}{*}{Barlow Twins} & Default & \pmb{$52.81 \pm 0.33$} & \pmb{$26.23 \pm 0.12$} & \pmb{$58.68 \pm 1.2$} \\
 & Bounded & $51.83 \pm 0.16$ & $24.99 \pm 0.38$ & $57.38 \pm 0.78$ \\ \hline
\end{tabular}
\end{table}

Table \ref{tbl:g_vs_f_results} shows the benefit of computing the affine loss using representations from $f$ instead of from projection head $g$. For both SimCLR and Barlow Twins, performance increases in all cases when using $f$ instead of $g$. We posit this is because this decouples the learned invariance from the learned sensitivity. Interestingly, the opposite is true for BYOL, and we attribute this to the extra MLP \emph{predictor head} in the BYOL architecture introducing more complicated learning dynamics.

\begin{table}[!h]
\caption{\label{tbl:g_vs_f_results}Results for computing $l_{\texttt{affine}}$ using representations from backbone encoder $f$ or projection head $g$.}
\centering
\begin{tabular}{ll|rrr}
 &  & \multicolumn{1}{l}{CIFAR10} & \multicolumn{1}{l}{CIFAR100} & \multicolumn{1}{l}{Caltech101} \\ \hline
\multirow{2}{*}{SimCLR} & $f$ & \pmb{$55.1 \pm 0.2$} & \pmb{$28.6 \pm 0.08$} & \pmb{$62.34 \pm 0.4$} \\
 & $g$ & $53.4 \pm 0.29$ & $26.5 \pm 0.15$ & $60.36 \pm 0.65$ \\ \hline \hline
\multirow{2}{*}{BYOL} & $f$ & $52.29 \pm 0.23$ & $25.86 \pm 0.38$ & $58.45 \pm 0.89$ \\
 & $g$ & \pmb{$53.47 \pm 0.37$} & \pmb{$26.29 \pm 0.46$} & \pmb{$60.3 \pm 0.64$} \\ \hline \hline
\multirow{2}{*}{Barlow Twins} & $f$ & \pmb{$52.81 \pm 0.33$} & \pmb{$26.23 \pm 0.12$} & \pmb{$58.68 \pm 1.2$} \\
 & $g$ & $50.75 \pm 0.29$ & $24.83 \pm 0.16$ & $57.45 \pm 0.61$ \\ \hline
\end{tabular}
\end{table}

\subsection{Transformation Component Analysis}
\begin{table*}[!t]
\caption{Ablation results for the different components that comprise an affine transformation (translation, shear, rotation, and scale).}
\label{tbl:transformation_component}
\centering
\begin{tabular}{ll|c|c|c|c|}
\cline{3-6}
 &  & Translation & Shear & Rotation & Scale \\ \hline
\multicolumn{1}{|l|}{SimCLR + Affine Module} & CIFAR10 & 54.14 & 54.77 & 54.74 & \pmb{55.41} \\ \hline
\multicolumn{1}{|l|}{SimCLR + Affine Module} & CIFAR100 & 28.12 & 28.78 & 27.71 & \pmb{28.88} \\ \hline
\multicolumn{1}{|l|}{SimCLR + Affine Module} & Caltech101 & 59.28 & \pmb{60.09} & 59.42 & 59.55 \\ \hline \hline
\multicolumn{1}{|l|}{BYOL + Affine Module} & CIFAR10 & \pmb{49.36} & 47.33 & 48.4 & 48.01 \\ \hline
\multicolumn{1}{|l|}{BYOL + Affine Module} & CIFAR100 & \pmb{24.62} & 22.04 & 23.67 & 24.03 \\ \hline
\multicolumn{1}{|l|}{BYOL + Affine Module} & Caltech101 & \pmb{54.24} & 52.09 & 52.94 & 52.15 \\ \hline \hline
\multicolumn{1}{|l|}{Barlow Twins + Affine Module} & CIFAR10 & 50.91 & 50.76 & \pmb{51.85} & 50.94 \\ \hline
\multicolumn{1}{|l|}{Barlow Twins + Affine Module} & CIFAR100 & 24.63 & 24.86 & \pmb{25.19} & 24.96 \\ \hline
\multicolumn{1}{|l|}{Barlow Twins + Affine Module} & Caltech101 & 55.55 & 54.88 & \pmb{55.68} & 54.74 \\ \hline
\end{tabular}
\end{table*}

Table \ref{tbl:transformation_component} shows the effect each of the four components of an affine transformation has on performance. In general, predicting the scale $\sigma$, rotation angle $\theta$, and translation components $t_x, t_y$ seems to have the largest effect on downstream performance for SimCLR, BYOL, and Barlow Twins, respectively. Predicting the scale $s_x, s_y$ is in general never the best-performing component to estimate. In almost all cases, simply predicting a single transformation component in isolation is sufficient to outperform the standard baseline SSL model without our affine module (refer to Table \ref{tbl:main_results}). Further, from Table \ref{tbl:percent_of_max_component_results} we see that BYOL benefits the most from training on all four constituent transformations since the accuracy drops the most significantly when training on the individual base transformations in isolation (it only retains approximately 92\% of the max accuracy). Conversely, SimCLR is the most robust to being trained on any of the individual constituent transformations in isolation. For example, when estimating the scale in isolation, an average of 99\% of the max accuracy is retained with SimCLR. These results suggest that, for all models, it is important to estimate the parameters of all four base transformations since the accuracy improves in all cases when doing so. Further, estimating the parameters of an additional base transformation imposes negligible additional computational cost - the dimensionality of the domain of the affine parameter estimator $r$ increases by 1 or 2.

\begin{table}[!h]
\caption{Percentage of maximum accuracy (defined by the accuracy when training on all four components as reported in Table \ref{tbl:main_results}) for the individual base transformations (mean over the downstream datasets).}
\label{tbl:percent_of_max_component_results}
\centering
\begin{tabular}{l|c|c|c|c|}
\cline{2-5}
 & \multicolumn{1}{l|}{Translation} & \multicolumn{1}{l|}{Shear} & \multicolumn{1}{l|}{Rotation} & \multicolumn{1}{l|}{Scale} \\ \hline
\multicolumn{1}{|l|}{SimCLR + Affine} & 97.22\% & 98.81\% & 97.18\% & 99.02\% \\ \hline
\multicolumn{1}{|l|}{BYOL + Affine} & 94.13\% & 88.29\% & 91.56\% & 91.32\% \\ \hline
\multicolumn{1}{|l|}{Barlow Twins + Affine} & 94.99\% & 94.81\% & 96.37\% & 94.97\% \\ \hline
\end{tabular}
\end{table}

\section{Conclusion}
\label{sec:conclusion}
We propose a module that performs affine transformation estimation as a way of improving the downstream performance and convergence rate of modern multi-view-based SSL algorithms. Through extensive experiments and ablations, we show that the models perform well in general across all studied models, namely, SimCLR, Barlow Twins, and BYOL. We run ablation studies on the core architectural components of our proposed model, including the aggregation scheme, the number of views from which to compute the affine loss, the potential bias introduced by the extra background induced from performing geometric transformations on images, and lastly the impact of the constituent transformation components. We find the optimal variant employs vector difference for aggregation and a single random view to estimate the affine loss. This parameterisation offers a good balance of downstream performance and additional computational burden.

The fact that the inclusion of our module results in notably improved downstream performance for all SSL models and downstream datasets suggests that the supervision signal derived from the affine module's loss contains complementary information not present within the standard ID loss and that SSL models benefit from a balance of invariance and sensitivity to different transformations.

\bibliographystyle{unsrtnat}
\bibliography{refs}

\end{document}